%% file: main.tex
\documentclass[10pt,twocolumn,letterpaper]{article}

\usepackage{cvpr}
\usepackage{times}
\usepackage{epsfig}
\usepackage{graphicx}
\usepackage{amsmath}
\usepackage{amssymb}
\usepackage{xcolor}

\usepackage{flushend}
\usepackage{booktabs}

\usepackage[pagebackref=true,breaklinks=true,letterpaper=true,colorlinks,bookmarks=false]{hyperref}

\graphicspath{{./img/}}

\newcommand{\MethodName}{gDLS*}

\newcommand{\fp}[1]{{\color{blue} \bf #1}} 
\newcommand{\zp}[1]{{\color{black} #1}} 

\cvprfinalcopy 

\def\cvprPaperID{5830} 

\ifcvprfinal\fi
\begin{document}

\title{\MethodName: Generalized Pose-and-Scale Estimation Given Scale and Gravity Priors}

\author{Victor Fragoso \\
Microsoft\\
{\tt\small victor.fragoso@microsoft.com}
\and
Joseph DeGol\\
Microsoft\\
{\tt\small joseph.degol@microsoft.com}
\and
Gang Hua$^{1}$\\
Wormpex AI\\
{\tt\small ganghua@gmail.com}
}

\maketitle




\begin{abstract}
Many real-world applications in augmented reality (AR), 3D mapping, and robotics require both fast and accurate estimation of camera poses and scales from multiple images captured by multiple cameras or a single moving camera. Achieving high speed and maintaining high accuracy in a pose-and-scale estimator are often conflicting goals. To simultaneously achieve both, we exploit {\em a priori} knowledge about the solution space. We present \MethodName, a generalized-camera-model pose-and-scale estimator that utilizes rotation and scale priors. \MethodName~ allows an application to flexibly weigh the contribution of each prior, which is important since priors often come from noisy sensors. Compared to state-of-the-art generalized-pose-and-scale estimators (\eg gDLS), our experiments on both synthetic and real data consistently demonstrate that \MethodName~ accelerates the estimation process and improves scale and pose accuracy.
\end{abstract}


\input{./tex/Introduction}
\input{./tex/RelatedWork}
\input{./tex/Method}
\input{./tex/Results}
\input{./tex/Conclusion}







\section*{Acknowledgment}
Gang Hua was supported in part by the National Key R\&D Program of China Grant 2018AAA0101400 and NSFC Grant 61629301.


\appendix
\section{Incorporating Priors via Regularizers}
\label{sec:derivation}

Section 3 of the main submission describes the proposed formulation to include scale and rotation priors using regularizers. This formulation is:

\begin{equation}
\begin{split}
    J^\prime &= J(R, \mathbf{t}, s, \boldsymbol{\alpha}) + \underbrace{\lambda_s \left( s_0 - s \right)^2}_{J_s} + \underbrace{\lambda_g \| \mathbf{g}_\mathcal{Q} \times R \mathbf{g}_\mathcal{W} \|^2}_{J_g},
    \label{eq:regularized_scale_and_pose_problem}
\end{split}
\end{equation}
where 
\begin{equation}
J(R, \mathbf{t}, s, \boldsymbol{\alpha}) = \sum_{i=1}^n \| \alpha_i \mathbf{r}_i - \left(R \mathbf{p}_i + \mathbf{t} - s \mathbf{c}_i \right)\|^2
    \label{eq:scale_and_pose_problem}
\end{equation}
is the gDLS~\cite{sweeney2014gdls} pose-and-scale re-projection error cost function from the $n$ 2D-3D correspondences, $J_s$ is the scale prior regularizer, and $J_g$ is the gravity direction constraint imposing a rotation prior. 

\subsection{Cost Function Depending Only on Rotation}
\label{sec:cost_rotation}
As mentioned in Section 3, the first step to obtain a cost function that we can minimize in a single shot is to rewrite Eq.~\eqref{eq:regularized_scale_and_pose_problem} as a function of the rotation matrix $R$. To do so, we define
\begin{equation}
\mathbf{x} = \begin{bmatrix} \alpha_1 & \hdots & \alpha_n & s &\mathbf{t}^\intercal\end{bmatrix}^\intercal,
\end{equation}
a vector holding the depths for each $i$-th point $\alpha_i$, the scale $s$, and translation vector $\mathbf{t}$. We know that the optimal depths, translation, and scale $\mathbf{x}^\star$ vanish the gradient $\nabla_{\mathbf{x}} J^\prime$ of the cost function $J^\prime$, \ie,
\begin{equation}
    \nabla_{\mathbf{x}} J^\prime \bigr |{_{\mathbf{x} = \mathbf{x}^\star}} = \left[ \nabla_{\mathbf{x}} J + \nabla_{\mathbf{x}} J_s \right]_{\mathbf{x} = \mathbf{x}^\star} = 0.
    \label{eq:partial}
\end{equation}
To satisfy this constraint, we calculate each of the gradients $\nabla_{\mathbf{x}} J$ and $\nabla_{\mathbf{x}} J_s$:
\begin{equation}
\begin{split}
    \nabla_{\mathbf{x}} J &=  2A^\intercal A \mathbf{x} - 2 A W b\\
    \nabla_{\mathbf{x}} J_s &= 2 P \mathbf{x} - 2 P \mathbf{x}_0,
\end{split}
\label{eq:partial_parts}
\end{equation}
where 
\begin{equation}
\begin{split}
A &= \begin{bmatrix}  
\mathbf{r}_1 & & & \mathbf{c}_1 & -I \\
 & \ddots & & \vdots & \vdots \\
 & & \mathbf{r}_n & \mathbf{c}_n & -I 
\end{bmatrix}, 
\mathbf{b} = \begin{bmatrix} \mathbf{p}_1 \\ \vdots \\ \mathbf{p}_n \end{bmatrix} \\
P &= \begin{bmatrix} 
  0_{n \times n} & & \\ 
  & \lambda_s & \\
  & & 0_{3 \times 3}
\end{bmatrix},
W = \begin{bmatrix} 
 R & & \\
 & \ddots & \\
 & & R
\end{bmatrix}, 
\end{split}
\end{equation}
and $\mathbf{x}_0 = \begin{bmatrix} 0_n^\intercal ~ s_o ~ 0_3^\intercal \end{bmatrix}^\intercal$. Note that 
\begin{equation}
J_s = \lambda_s \left( s_0 - s\right)^2 = \left( \mathbf{x} - \mathbf{x}_0 \right)^\intercal P \left( \mathbf{x} - \mathbf{x}_0 \right).
\end{equation}
Thus, we can rewrite Eq.~\eqref{eq:partial} into
\begin{equation}
    \nabla_{\mathbf{x}} J^\prime = A^\intercal A \mathbf{x} - A W b + P \mathbf{x} - P \mathbf{x}_0 = 0,
    \label{eq:partial_simplified}
\end{equation}
by combining Equations~\eqref{eq:partial} and ~\eqref{eq:partial_parts}. Rearranging terms of Eq.~\eqref{eq:partial_simplified} yields Eq. (5) in the main submission, which is
\begin{equation}
    \mathbf{x} = \left(A^\intercal A  + P\right)^{-1} A^\intercal Wb + \left(A^\intercal A + P \right)^{-1} P\mathbf{x}_0.
    \label{eq:concrete_partial}
\end{equation}

In order to obtain the simplified version of Eq.~\eqref{eq:concrete_partial} shown in Eq. (5) of the main submission and inspired by~\cite{sweeney2014gdls}, we rewrite
\begin{equation}
\begin{split}
    A^\intercal A + P &= \begin{bmatrix} 
    I_{n \times n} & B^\prime_{n \times 4} \\ B^{\prime \intercal}_{n \times 4} & D^\prime_{4 \times 4}
    \end{bmatrix} \\
    &= \underbrace{\begin{bmatrix} 
    I_{n \times n} & B_{n \times 4} \\ B^\intercal_{n \times 4} & D_{4 \times 4}
    \end{bmatrix}}_{A^\intercal A} + \underbrace{\begin{bmatrix} 
    0_{n \times n} & & \\ 
    & \lambda_s & \\
    & & 0_{3 \times 3}
    \end{bmatrix}}_{P},
\end{split}
\end{equation}
where 
\begin{equation}
\begin{split}
    B &= \begin{bmatrix} 
    \mathbf{r}_1^\intercal \mathbf{c}_1 & -\mathbf{r}_1^\intercal \\
    \vdots & \vdots \\
    \mathbf{r}_n^\intercal \mathbf{c}_n & -\mathbf{r}_n^\intercal \\
    \end{bmatrix} \\
    D &= \begin{bmatrix} 
    \sum_{i=1}^n \mathbf{c}_i^\intercal \mathbf{c}_i & \sum_{i=1}^n -\mathbf{c}^\intercal_i \\
    \sum_{i=1}^n -\mathbf{c}_i & nI
    \end{bmatrix},
\end{split}
\end{equation}
and $I$ is the identity matrix. It is important to mention that $A^\intercal A$ is exactly the same as that of gDLS~\cite{sweeney2014gdls}, and thus
\begin{equation}
\begin{split}
    A^\intercal A + P &= \begin{bmatrix} 
    I_{n \times n} & B_{n \times 4} \\ B^\intercal_{n \times 4} & D^\prime_{4 \times 4}
    \end{bmatrix} \\
    D^\prime_{4 \times 4} &= \begin{bmatrix} 
    \lambda_s + \sum_{i=1}^n \mathbf{c}_i^\intercal \mathbf{c}_i & \sum_{i=1}^n -\mathbf{c}^\intercal_i \\
    \sum_{i=1}^n -\mathbf{c}_i & nI
    \end{bmatrix}.
\end{split}
\end{equation}

Eq.~\eqref{eq:concrete_partial} requires the inverse of $\left(A^\intercal A + P\right)^{-1}$. To compute a closed form relationship, we use the following block matrix expression
\begin{equation}
    \left(A^\intercal A + P\right)^{-1} = \begin{bmatrix} E_{n \times n} ~ F_{n \times 4} \\ G_{4 \times n} ~ H_{4 \times 4} \end{bmatrix}.
\end{equation}
Through block matrix inversion, we obtain the following closed-form block matrices:
\begin{equation}
    \begin{split}
    E &= I + B H B^\intercal \\
    F &= -B H \\
    G &= -H B^\intercal \\
    H &= \left( D^\prime - Y \right)^{-1} \\
    Y &=  \begin{bmatrix} \sum_{i=1}^n \mathbf{c}_i^\intercal \mathbf{r}_i \mathbf{r}_i^\intercal \mathbf{c}_i & \sum_{i=1}^n -\mathbf{c}_i^\intercal \mathbf{r}_i \mathbf{r}_i^\intercal \\ \sum_{i=1}^n -\mathbf{r}_i \mathbf{r}_i^\intercal \mathbf{c}_i & \sum_{i=1}^n \mathbf{r}_i \mathbf{r}_i^\intercal \end{bmatrix} \\
    \end{split}
\end{equation}

Like in gDLS~\cite{sweeney2014gdls}, we use matrices $U$, $S$, and $V$ to simplify Eq.~\eqref{eq:concrete_partial}, \ie,
\begin{equation}
    (A^\intercal A + P)^{-1} A^\intercal = \begin{bmatrix}
    U \\ S \\ V
    \end{bmatrix},
    \label{eq:usv_mats}
\end{equation}
where
\begin{equation}
\begin{split}
    U &= \begin{bmatrix} \mathbf{r}_1^\intercal & & \\ & \ddots & \\ & & \mathbf{r}_n^\intercal \end{bmatrix} + B \begin{bmatrix} S \\ V\end{bmatrix} \\
    \begin{bmatrix} S \\ V\end{bmatrix} &= -H B^\intercal \begin{bmatrix} \mathbf{r}_1^\intercal & & \\ & \ddots & \\ & & \mathbf{r}_n^\intercal \end{bmatrix} + H \begin{bmatrix} \mathbf{c}_1^\intercal & \hdots & \mathbf{c}_n^\intercal \\
    -I & \hdots & -I \end{bmatrix} \\
     &= H \begin{bmatrix} \mathbf{c}_1^\intercal - \mathbf{c}_1 \mathbf{r}_1 \mathbf{r}_1^\intercal & \hdots & \mathbf{c}_n^\intercal - \mathbf{c}_n \mathbf{r}_n \mathbf{r}_n^\intercal \\
     \mathbf{c}_1 \mathbf{c}_1^\intercal - I & \hdots & \mathbf{c}_n \mathbf{c}_n^\intercal - I \end{bmatrix}.
\end{split}
\end{equation}

We can simplify Eq.~\eqref{eq:concrete_partial} further. To do this, we focus on simplifying the term encoding the scale prior, yielding
\begin{equation}
\begin{split}
    (A^\intercal A + P)^{-1} P \mathbf{x}_0 &= \begin{bmatrix} 
    E & F \\ G & H
    \end{bmatrix} 
    \underbrace{\begin{bmatrix} 
    0 & & \\
      & \lambda_s & \\
      & & 0
    \end{bmatrix}}_{P}
    \underbrace{\begin{bmatrix} 
    0 \\ s_0 \\ 0
    \end{bmatrix}}_{\mathbf{x}_0} \\
    &= \lambda_s s_0 \underbrace{\begin{bmatrix} F_1 \\ B_1 \end{bmatrix}}_{\mathbf{l}}
\end{split},
\label{eq:scale_prior_term}
\end{equation}
where $F_1$ and $B_1$ are the first column of the matrix $F$ and $B$, respectively. Combining Equations~\eqref{eq:scale_prior_term} and~\eqref{eq:usv_mats} allows us to rewrite Eq.~\eqref{eq:concrete_partial} as follows:
\begin{equation}
    \mathbf{x} = \begin{bmatrix} U \\ S \\ V \end{bmatrix} W \mathbf{b} + \lambda_s s_0 \mathbf{l},
    \label{eq:linear_relationship}
\end{equation}
which is the bottom part of Eq. (5) in the main submission. Eq.~\eqref{eq:linear_relationship} provides a linear relationship between depths, scale, and translation and the rotation matrix. The explicit relationships are the following

\begin{equation}
\begin{split}
\alpha_i(R) &= \mathbf{u}_i^\intercal W \mathbf{b} + \lambda_s s_o \mathbf{l}_i \\
s(R) &= S W \mathbf{b} + \lambda_s s_o \mathbf{l}_{n + 1} \\
\mathbf{t}(R) &= V W \mathbf{b} + \lambda_s s_o \mathbf{l}_{\mathbf{t}},
\end{split}
\label{eq:rotation_relationships}
\end{equation}
where $\mathbf{u}_i^\intercal$ is the $i$-th row of matrix $U$, $\mathbf{l}_j$ is the $j$-th entry of the vector $\mathbf{l}$, and $\mathbf{l}_{\mathbf{t}}$ corresponds to the last three entries of the vector $\mathbf{l}$. Specifically, the entries of vector $\mathbf{l}$ are

\begin{equation}
    \mathbf{l} = \begin{bmatrix} \mathbf{l}_1 \\ \vdots \\ \mathbf{l}_n \\ \mathbf{l}_{n + 1} \\ \mathbf{l}_{\mathbf{t}}\end{bmatrix} = \begin{bmatrix} F_{1, 1} \\ \vdots \\ F_{n, 1} \\ H_{1, 1}  \\ H_{2:4, 1} \end{bmatrix},
    \label{eq:vector_l}
\end{equation}
where $H_{2:4, 1}$ represent the last three entries of the first column of $H$. We can use these explicit relationships (\ie, Eq.~\eqref{eq:regularized_scale_and_pose_problem} and Eq.~\eqref{eq:vector_l}) to rewrite the main cost function as one depending only on rotation parameters. To do so as clearly as possible, we define 

\begin{equation}
\begin{split}
\mathbf{e}_i &= \alpha_i(R) \mathbf{r}_i - \left(R \mathbf{p}_i + \mathbf{t}(R) - s(R)\mathbf{c}_i \right) \\
 &= \left( \mathbf{u}_i^\intercal W \mathbf{b} + \lambda s_0 F_{i,1} \right)\mathbf{r}_i -R\mathbf{p}_i \\
 &~~~ - \left(VW\mathbf{b} + \lambda s_0 H_{2:4, 1}\right) \\
 &~~~ +\left(SW\mathbf{b} + \lambda s_0 H_{1,1} \right)\mathbf{c}_i \\
 & = \underbrace{\mathbf{u}_i^\intercal W \mathbf{b}\mathbf{r}_i -R\mathbf{p}_i -VW\mathbf{b} + SW\mathbf{b}\mathbf{c}_i}_{\boldsymbol{\eta}_i} \\
 &~~~ + \underbrace{\lambda s_0 \left( F_{i,1}\mathbf{r}_i - H_{2:4,1} + H_{1,1}\mathbf{q}_i\right)}_{\mathbf{k}_i} \\
 & = \boldsymbol{\eta}_i + \mathbf{k}_i
\end{split}.
\end{equation}

As noted in gDLS~\cite{sweeney2014gdls} paper, $\boldsymbol{\eta}_i$ can be factored out as follows:
\begin{equation}
\begin{split}
\boldsymbol{\eta}_i &= \left(\mathbf{r}_i \mathbf{r}_i^\intercal - I \right)\left(R\mathbf{p}_i -SW \mathbf{b} \mathbf{c}_i + VW \mathbf{b} \right) \\
& = \underbrace{\left(\mathbf{r}_i\mathbf{r}_i^\intercal - I \right)\begin{bmatrix}L(\mathbf{p}_i) & -\mathbf{c}_i SL(\mathbf{b}) & VL(\mathbf{b}) \end{bmatrix}}_{M_i}\text{vec}(R) \\
\end{split},
\label{eq:eta_factorization}
\end{equation}
where $\text{vec}(R)$ vectorizes a rotation matrix R, and $L(\mathbf{z})$ is a function that computes a matrix such that $R\mathbf{z} = L(\mathbf{z}) \text{vec}(R)$. Since we use the rotation representation of Upnp~\cite{kneip2014upnp}, \ie, 
\begin{equation}
\text{vec}(R) = \left[
q_1^2 ~ q_2^2 ~ q_3^2 ~ q_4^2 ~ q_1 q_2 ~ q_1 q_3 ~ q_1 q_4 ~ q_2 q_3 ~ q_2 q_4 ~ q_3 q_4 \right]^\intercal,
\end{equation}
then the function $L(\cdot)$ is
\begin{equation}
    L(\mathbf{z})^\intercal = \begin{bmatrix} 
    \mathbf{z}_1 & \mathbf{z}_2 & \mathbf{z}_3 \\
    \mathbf{z}_1 & -\mathbf{z}_2 & -\mathbf{z}_3 \\
    -\mathbf{z}_1 & \mathbf{z}_2 & -\mathbf{z}_3 \\
    -\mathbf{z}_1 & -\mathbf{z}_2 & \mathbf{z}_3 \\
    0 & -2\mathbf{z}_3 & 2\mathbf{z}_2 \\
    2\mathbf{z}_3 & 0 & -2\mathbf{z}_1 \\
    -2\mathbf{z}_2 & 2\mathbf{z}_1 & 0 \\
    2\mathbf{z}_2 & 2\mathbf{z}_1 & 0 \\
    2\mathbf{z}_3 & 0 & 2\mathbf{z}_1 \\
    0 & 2\mathbf{z}_3 & 2\mathbf{z}_2
    \end{bmatrix}.
\end{equation}

By substituting the relationships shown in Eq.~\eqref{eq:rotation_relationships} and the factorizations shown in Eq.~\eqref{eq:eta_factorization} into Eq.~\eqref{eq:regularized_scale_and_pose_problem}, we obtain the following relationships:
\begin{equation}
\begin{split}
J^\prime_{\text{gDLS}} &= \sum_{i=1}^n \mathbf{e}_i^\intercal \mathbf{e}_i = \sum_{i=1}^n \boldsymbol{\eta}_i^\intercal \boldsymbol{\eta}_i + 2 \mathbf{k}_i^\intercal \boldsymbol{\eta}_i + \mathbf{k}_i^\intercal \mathbf{k}_i \\
 &= \sum_{i=1}^n \text{vec}(R)^\intercal M_i^\intercal M_i  \text{vec}(R) + 2\mathbf{k}_i^\intercal M_i \text{vec}(R) + \mathbf{k}_i^\intercal \mathbf{k}_i \\
 &= \text{vec}(R)^\intercal \underbrace{\left(\sum_{i=1}^n M_i^\intercal M_i \right)}_{M_{\text{gDLS}}} \text{vec}(R) + \\
  & ~~~~~~ 2\underbrace{\left(\sum_{i=1}^n \mathbf{k}_i^\intercal M_i \right)}_{\mathbf{d}_{\text{gDLS}}^\intercal}\text{vec}(R) + \underbrace{\sum_{i=1}^n \mathbf{k}_i^\intercal \mathbf{k}_i}_{k_{\text{gDLS}}} \\
 &= \text{vec}(R)^\intercal M_{\text{gDLS}} \text{vec}(R)  + 2\mathbf{d}^\intercal_{\text{gDLS}} \text{vec}(R) + k_{\text{gDLS}} \\
\end{split};
\label{eq:unconstrained_gdls}
\end{equation}

\begin{equation}
\begin{split}
 J^\prime_s &= \lambda_s \left(s_0 - s(R) \right)^2 \\
  &= \lambda_s \left(SL(\mathbf{b}) \text{vec}(R) + \lambda_s s_0 H_{1,1} - s_0 \right)^2 \\
  &= \text{vec}(R)^\intercal \underbrace{\left(\lambda_s L(\mathbf{b})^\intercal S^\intercal S L(\mathbf{b}) \right)}_{M_s} \text{vec}(R) + \\
  & ~~~~~ 2 \underbrace{\lambda_s \left(s_0 - \lambda_s s_0 H_{1,1}\right) SL(\mathbf{b})}_{\mathbf{d}_s^{\intercal}} \text{vec}(R)  + \\
  & ~~~~~ \underbrace{\lambda_s \left( \lambda_s s_0 H_{1,1} - s_0 \right)^2}_{k_s} \\
  &= \text{vec}(R)^\intercal M_s \text{vec}(R) + 2 \mathbf{d}_s^\intercal \text{vec}(R) + k_s
\end{split}; \text{and}
\label{eq:scale_regularizer}
\end{equation}

\begin{equation}
\begin{split}
 J^\prime_g &= \lambda_g \| \mathbf{g}_\mathcal{Q} \times R \mathbf{g}_\mathcal{W}\|^2 \\
  &= \text{vec}(R)^\intercal \underbrace{\left(\lambda_g  L(\mathbf{g}_{\mathcal{W}})^\intercal \lfloor \mathbf{g}_{\mathcal{Q}} \rfloor_\times^\intercal \lfloor \mathbf{g}_{\mathcal{Q}} \rfloor_\times L(\mathbf{g}_{\mathcal{W}}) \right)}_{M_g} \text{vec}(R) \\
  &= \text{vec}(R)^\intercal M_g \text{vec}(R)
\end{split}.
\label{eq:gravity_regularizer}
\end{equation}
The symbol $\lfloor \cdot \rfloor_\times$ indicates the skew symmetric matrix. By putting together the components of the cost, we end up with the final cost function
\begin{equation}
    J^\prime = J^\prime_{\text{gDLS}} + J^\prime_s  + J^\prime_g,
\end{equation}
which is Eq. (10) in the main submission.


{\small
\bibliographystyle{ieee_fullname}
\bibliography{references}
}


\end{document}

%% file: tex/Introduction.tex
\vspace{-2mm}
\section{Introduction}
\label{sec:introduction}

\begin{figure}[t]
    \centering
    \includegraphics[width=0.47\textwidth]{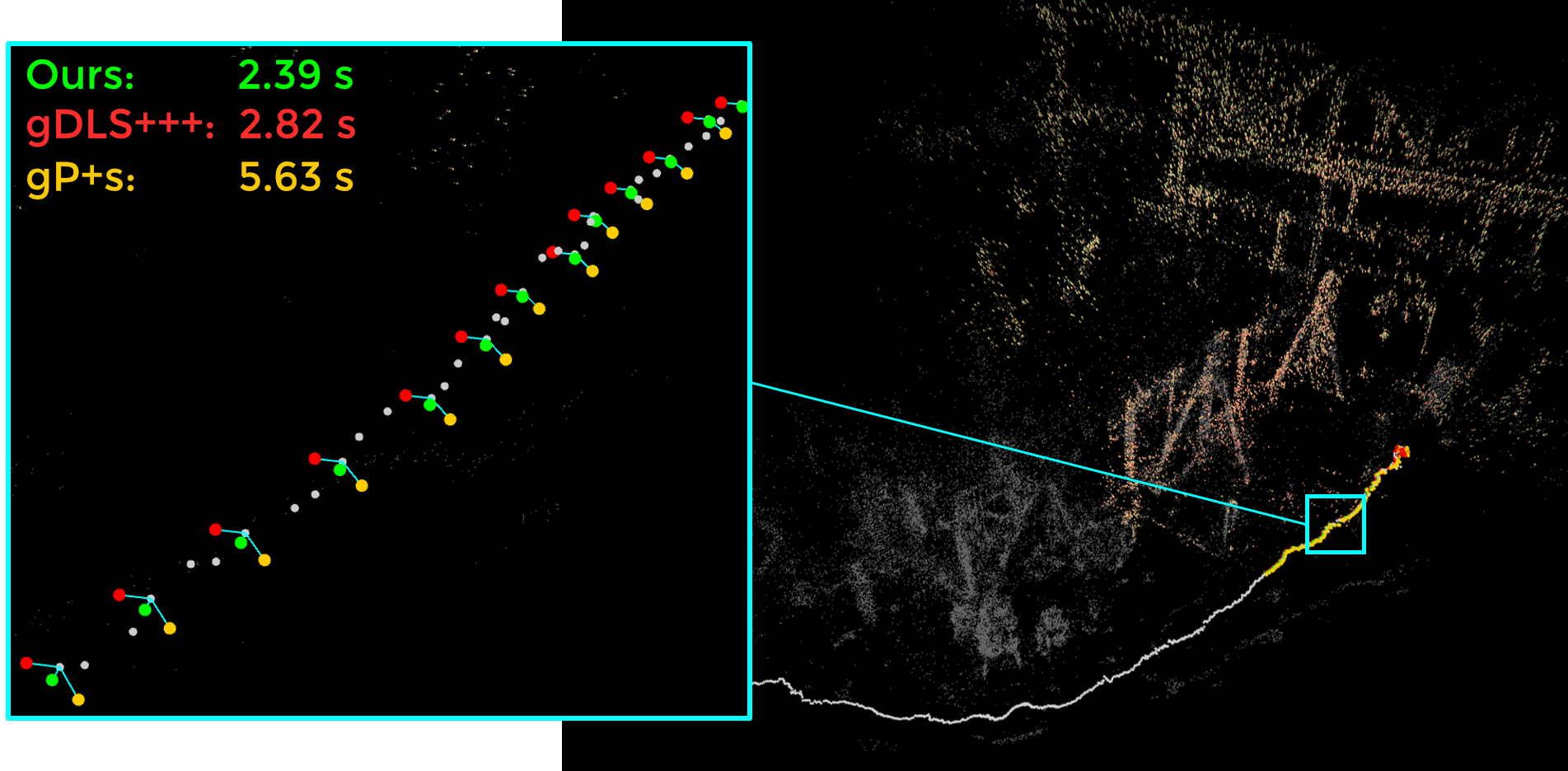}
    \caption{Estimating the pose and scale accurately and quickly is essential in many applications in AR, 3D mapping, and robotics. We introduce \MethodName, a pose-and-scale estimator that exploits scale and/or gravity priors to improve accuracy and speed. Compared to state-of-the-art estimators (\eg, gDLS+++~\cite{sweeney2016large} and gP+s~\cite{ventura2014minimal}), \MethodName~ achieves more accurate estimates in less time when registering a set of cameras to an existing 3D reconstruction. The right image above shows an existing 3D reconstruction (gray) with aligned cameras and points for \MethodName~(green), gDLS+++~(red), and gP+s~(orange). The left image shows a zoomed view of the positions of the aligned cameras. The white points are the expected camera positions and the green/red/orange points are the estimate positions for each method. The cyan line between indicates a match, where longer lines indicate more error.}
    \label{fig:intro_fig}
\end{figure}

\let\thefootnote\relax\footnote{$^{1}$ This work was done while at Microsoft.}
\let\thefootnote\relax\footnote{$^{2}$ \url{josephdegol.com/pages/GDLSStar_CVPR20.html}}
\let\thefootnote\relax\footnote{Published at CVPR 2020.}
Estimating the pose and scale from multiple images taken from multiple cameras or multiple images from one moving camera (\eg, a SLAM~\cite{engel2014lsd, jones2011visual, klein2009parallel, mur2017orb, tsotsosCS15} trajectory) is an essential step in many augmented reality (AR)~\cite{martin2014mapping, schmalstieg2016augmented, sweeney2015efficient, ventura2012wide, ventura2014global}, 3D mapping~\cite{camposeco2018hybrid, geppert2018efficient, ma2012invitation, sattler2018benchmarking, sweeney2014solving, sweeney2016large}, and robotics applications~\cite{heng2018project, kneip2011robust, kneip2013using, Levinson2007MapBasedPV, weiss2013monocular, ziegler2014video}. Consider hologram sharing services (\eg, Azure Spatial Anchors~\cite{microsoft_asa}) as an example. These services have a reference map and need to localize query images accurately (so that holograms are positioned correctly) and quickly (to maintain a nice user experience). However, as Figure~\ref{fig:intro_fig} shows, current methods leave room for improvement in terms of both accuracy and processing time. In this work, we propose \MethodName, a multi-camera pose-and-scale estimator that exploits scale and gravity priors to improve accuracy and speed. Despite using additional information, \MethodName~ computes its parameters with linear complexity in the number of points and multiple optimal solutions in a single shot, avoiding iterative optimization procedures.

Using single camera pose estimators (\eg, \cite{bujnak2008general, Ferraz_2014_CVPR, hesch2011direct, kneip2011novel, kukelova2013real, lepetit2009epnp, miraldo_icra14, Zheng_2013_ICCV}) to develop a multi-camera pose-and-scale estimator is cumbersome, and their estimates tend to be inaccurate~\cite{kneip2014upnp, sweeney2014gdls}. Instead, many multi-camera pose-and-scale estimators~\cite{kneip2014upnp, sweeney2014gdls, sweeney2016large} use the generalized camera model~\cite{grossberg_iccv19,pless2003using} to elegantly treat the collection of cameras as one generalized camera, yielding accuracy improvements. Despite their improvements, these estimators often produce erroneous results due to noisy input data and numerical instabilities in their underlying polynomial solvers.


Given the need of accurate pose and scale estimates by many applications in AR, 3D mapping, and robotics, some algorithms~\cite{sweeney2015efficient, sweeney2014solving, zeisl2015camera} exploit inertial measurements (\eg, gravity directions). Most of these approaches assume that the gravity or down directions are reliable, and include this extra knowledge as part of their mathematical derivation to simplify the problem. However, the gravity directions can still be noisy due to the nature of these sensors and can affect the accuracy of the estimates. In contrast, \MethodName~ adopts a generalized-camera model with regularizers that encode scale and rotation priors (\eg, gravity direction). These regularizers allow a user to independently control the contribution of each individual prior, which is beneficial to reduce the effect of noise present in each prior.

We show using synthetic data that \MethodName~ is numerically stable and resilient to noise. We demonstrate this by (1) varying pixel noise and sample size and showing that \MethodName~ estimates transformations with errors that are no worse than current estimators; and (2) varying the noise in the scale and gravity priors and showing that \MethodName~maintains accuracy and speed. We then use real data (\ie~\cite{Geiger2013IJRR,sturm12iros}) to evaluate \MethodName~when registering a set of cameras to an existing 3D reconstruction. Our extensive experiments show that \MethodName~ is significantly faster and slightly more accurate than current pose-and-scale estimators (\ie,~\cite{kneip2014upnp,sweeney2014gdls, sweeney2016large, ventura2014minimal}). Moreover, the experiments show that a rotation prior based on gravity directions improves rotation and translation estimates while achieving significant speed-ups. On the other hand, a scale prior mainly improves scale estimates while modestly enhancing translation estimates and speed.

In summary, the \textbf{contributions} of this work are (1) \MethodName, a novel and generalized formulation of gDLS~\cite{sweeney2014gdls} that includes scale and gravity priors that computes its parameters with an $\mathcal{O}(n)$ complexity; (2) a novel evaluation protocol for pose-and-scale estimators that reports rotation, translation, and scale errors; and (3) extensive experimental results showing that \MethodName~consistently improves pose accuracy in less time.

%% file: tex/RelatedWork.tex
\section{Related Work}
\label{sec:related_work}

Estimating the position and orientation of a camera is crucial for many applications because they need to accurately register computer-generated content into the real-world, localize an agent (\eg, a visually impaired person) within an environment, and autonomously navigate (\eg, self-driving cars). While these applications use camera pose estimators to operate, most of the estimators have focused on localizing single cameras. Although these estimators~\cite{bujnak2008general, ferraz2014very, kneip2011novel, kukelova2010closed, kukelova2013real, lepetit2009epnp, wu2015p3, Zheng_2013_ICCV} have achieved impressive performance and accuracy, many applications~\cite{kneip2013using} have started to adopt multi-camera systems. This is because a multi-camera system can provide additional information that allows an application to estimate its pose more accurately. For this reason, this section reviews existing work on multi-camera pose and pose-and-scale estimators. 

\begin{table}[t]
    \centering
    \caption{\MethodName~ compares favorably to existing state-of-the-art pose-and-scale estimators because it maintains all the properties of other estimators while also being the only estimator that enables the use of gravity and scale priors.}
    {\setlength{\tabcolsep}{0.38em}
    \footnotesize {
    \begin{tabular}{l ccccc}
        \toprule
                  & gP+s & gDLS & gDLS+++ & UPnP & Ours\\
        Reference & \cite{ventura2014minimal} & \cite{sweeney2014gdls} & \cite{sweeney2016large} & \cite{kneip2014upnp} & -\\
        \cmidrule{1-6}
        Year & 2014 & 2014 & 2016 & 2014 & 2019\\
        Generalized Camera       & \checkmark & \checkmark & \checkmark & \checkmark & \checkmark\\
        Geometric Optimality      & \checkmark & \checkmark & \checkmark & \checkmark & \checkmark\\
        Linear Complexity         & & \checkmark & \checkmark & \checkmark & \checkmark\\
        Multiple Solutions        & \checkmark & \checkmark & \checkmark & \checkmark & \checkmark\\
        Similarity Transformation & \checkmark & \checkmark & \checkmark & & \checkmark \\
        Singularity-Free Rotation & \checkmark & & \checkmark & \checkmark & \checkmark\\
        Gravity Prior             & & & & & \checkmark\\
        Scale Prior               & & & & & \checkmark\\
        \bottomrule
    \end{tabular}
    }}
    \label{tab:relworks}
\end{table}

\subsection{Multi-Camera Pose Estimators}

Chen and Chang~\cite{chen_icra02} and Nister and Stewenius~\cite{nister2007minimal} proposed gP3P, a minimal estimator which requires three 2D-3D correspondences to estimate the pose of a multi-camera system. gP3P computes up to eight solutions by finding the intersections of a circle and a ruled quartic surface. Lee~\etal~\cite{HeeLee2016} also introduced a minimal estimator that utilizes Pl{\"u}cker lines to estimate the depth of each point. Subsequently, it estimates the position of each point \wrt to the frame of reference of the multi-camera system. Then it estimates the absolute pose of the multi-camera system.

Unlike previous minimal solvers, Kneip~\etal~\cite{kneip2014upnp} introduced UPnP, an efficient minimal and non-minimal pose estimator derived from a least-squares reprojection-error-based cost function. Inspired by DLS~\cite{hesch2011direct}, UPnP reformulates the cost function as one depending only on a unit-norm quaternion. UPnP finds the optimal rotation by solving a polynomial system that encodes the vanishing of the cost gradient at the optimal unit-norm quaternion via a Gr{\"o}bner-basis solver~\cite{kukelova2008automatic, larsson2018beyond}. 

\subsection{Pose-and-Scale Estimators}
Different from multi-camera pose estimators, pose-and-scale estimators compute the pose of a multi-camera system and a scale value; this value scales the positions of the cameras in order to align a 3D representation into the frame of reference of the multi-camera system more accurately.

Ventura~\etal~\cite{ventura2014minimal} proposed gP+s, a minimal pose-and-scale estimator that requires four 2D-3D correspondences and a Gr{\"o}bner basis polynomial solver~\cite{kukelova2008automatic, larsson2018beyond}. However, gP+s can also work with more than four points. Kukelova~\etal~\cite{kukelova2016efficient} introduced another minimal pose-and-scale estimator that avoids using a Gr{\"o}bner basis polynomial solver, leading to impressive speed-ups but a decrease in accuracy. 

Unlike previous estimators, Sweeney~\etal~\cite{sweeney2014gdls} presented gDLS, an estimator derived from a least-squares reprojection-error cost function. gDLS derives a cost function that depends only on a rotation matrix. Inspired by DLS~\cite{hesch2011direct}, gDLS solves a polynomial system that encodes the vanishing of the cost gradient at the optimal Cayley-Gibbs-Rodrigues angle-axis vector using the DLS polynomial solver (a Macaulay solver). Unfortunately, its Macaulay solver can be slow since it requires obtaining the eigenvectors of a $27 \times 27$ action matrix. To alleviate this issue, Sweeney~\etal~\cite{sweeney2016large} introduced gDLS+++, a gDLS-based estimator using a unit-norm quaternion. Thanks to the rotation representation of gDLS+++, it can use the efficient UPnP polynomial solver.


Different from previous methods, \MethodName~ is one of the first estimators to incorporate scale and rotation priors. As we show in Section~\ref{sec:results}, these priors improve both speed and accuracy. Moreover, \MethodName~ maintains many of the desirable properties of current solvers: (1) uses a generalized camera model which elegantly simplifies the formulation; (2) computes multiple optimal solutions in a single shot, avoiding iterative optimization procedures; (3) scales linearly when building its parameters; and (4) uses a singularity-free rotation representation. See Table~\ref{tab:relworks} for a brief comparison of estimator properties. 


%% file: tex/Method.tex
\section{Pose-and-Scale Estimation using Priors}
\label{sec:method}
The goal of \MethodName~ is to provide hints about the scale and rotation parameters of the similarity transformation using a generalized pose-and-scale estimator (\eg, gDLS~\cite{sweeney2014gdls}). Thanks to the prevalence of inertial sensors in mobile devices, these priors are readily available. For instance, a rotation prior can be obtained from the gravity direction using measurements from inertial sensors, and a scale prior can be obtained from the IMU~\cite{Nutzi2011}, GPS, or known landmark sizes~\cite{degol2018improved}. 

One of the design considerations of \MethodName~ is the ability to control the contribution of each of the priors independently. This allows the user to either disable or enable each of the priors. When enabling the priors, \MethodName~ allows a user to set a weight for each prior to control their confidence. In Section~\ref{sec:results}, we test a range of weights, but we plan to explore in future work how to set these weights automatically using the variance of the noise of the sensors. Because  \MethodName~ is based on the pose-and-scale formulations of gDLS~\cite{sweeney2014gdls, sweeney2016large}, we first describe the pose-and-scale formulation and then present our modifications that enable the use of scale and rotation priors.

\subsection{gDLS - A Pose-and-Scale Estimator Review}
\label{sec:gdls_formulation}

\begin{figure}[t]
\centering
\includegraphics[width=0.4\textwidth]{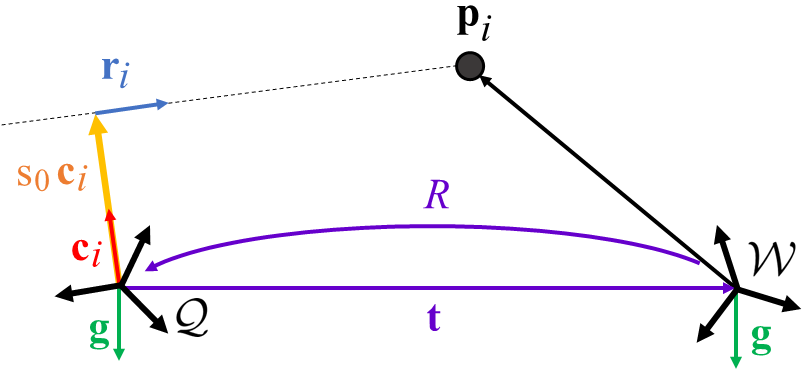}
\caption{Estimating the pose of a multi-camera system $\mathcal{Q}$ requires the estimation of $R$ and $\mathbf{t}$, while the scale $s$ adjusts the camera positions $\mathbf{c}_i$ so that $\mathcal{W}$ and $\mathcal{Q}$ use the same metric scale. \MethodName~ can use the gravity directions $\mathbf{g}$ to impose a rotation prior and a scale-prior $s_0$ to place the cameras at the right scale.}
\label{fig:eq1_figure}
\end{figure}

Given $n$ 2D-3D correspondences, gDLS computes the scale and pose of a non-central camera by minimizing the following least-squares cost function:
\begin{equation}
    J(R, \mathbf{t}, s, \boldsymbol{\alpha}) = \sum_{i=1}^n \| \alpha_i \mathbf{r}_i - \left(R \mathbf{p}_i + \mathbf{t} - s \mathbf{c}_i \right)\|^2,
    \label{eq:scale_and_pose_problem}
\end{equation}
where $\mathbf{r}_i$ is a unit-vector indicating the direction from the position of the camera $\mathbf{c}_i$ to a 3D point $\mathbf{p}_i$; $\alpha_i$ is the depth of the point $\mathbf{p}_i$ with respect to the camera position $\mathbf{c}_i$; $\boldsymbol{\alpha}$ is a vector holding the depths; $R \in \text{SO}(3)$ is a rotation matrix; $\mathbf{t} \in \mathbb{R}^3$ is a translation vector; and $s \in \mathbb{R}$ is the scale; see Fig.~\ref{fig:eq1_figure} for a visual representation of Eq.~\eqref{eq:scale_and_pose_problem}. 

The pose-and-scale formulation shown in Eq.~\eqref{eq:scale_and_pose_problem} accumulates the errors between the transformed $i$-th 3D point $\left(R \mathbf{p}_i + \mathbf{t} - s \mathbf{c}_i \right)$ and the same point described with respect to the camera $\alpha_i \mathbf{r}_i$. The rotation $R$, the translation $\mathbf{t}$, and $s \mathbf{c}_i$ transform a 3D point from a world coordinate system to the coordinate system of a generalized camera.

To find the minimizer $\left(R^\star, \mathbf{t}^\star, s^\star, \boldsymbol{\alpha}^\star \right)$, gDLS~\cite{sweeney2014gdls} first rewrites $J(R, \mathbf{t}, s, \boldsymbol{\alpha
})$ as a function that only depends on the rotation matrix. As Hesch and Roumeliotis~\cite{hesch2011direct} and Sweeney~\etal~\cite{sweeney2014gdls, sweeney2016large} demonstrated, the translation $\mathbf{t}$, scale $s$, and depth $\alpha_i$ can be written as a linear function of the rotation matrix $R$. Thus, it is possible to re-write the pose-and-scale least-squares cost formulation as follows:
\begin{equation}
\begin{split}
    J(R) &= \sum_{i=1}^n \| \alpha_i(R) \mathbf{r}_i - \left(R \mathbf{p}_i + \mathbf{t}(R) - s(R) \mathbf{c}_i \right)\|^2 \\
     &= \text{vec}(R)^\intercal M \text{vec}(R)
    \label{eq:quadratic_scale_and_pose_problem},
\end{split}
\end{equation}
where $\text{vec}(R)$ is a vectorized form of the rotation matrix, and $M$ is a square matrix capturing the constraints  from the input 2D-3D correspondences; the dimensions of $M$ depend on the vectorization and representation of $\text{vec}(R)$.

Given the cost function $J(R)$, gDLS finds the optimal rotation $R^\star$ by solving a polynomial system representing the constraint that the gradient $\nabla_{\mathbf{q}} J(R^\star) = 0$ is null with respect to the rotation parameters $\mathbf{q}$, and rotation-parameter constraints (\eg, ensuring a unit-norm quaternion). 

\subsection{Incorporating Priors via Regularizers}

In order to impose scale and rotation priors to Eq.~\eqref{eq:scale_and_pose_problem}, \MethodName~ uses regularizers. Adding these regularizers leads to the following least-squares cost function:
\begin{equation}
\begin{split}
    J^\prime &= J(R, \mathbf{t}, s, \boldsymbol{\alpha}) + \lambda_s \left( s_0 - s \right)^2 + \lambda_g \| \mathbf{g}_\mathcal{Q} \times R \mathbf{g}_\mathcal{W} \|^2,
    \label{eq:regularized_scale_and_pose_problem}
\end{split}
\end{equation}
where $s_0$ is the scale prior; $\mathbf{g}_\mathcal{Q}$ and $\mathbf{g}_\mathcal{W}$ are the gravity directions of the multi-camera setting and world, respectively; the symbol $\times$ represents the cross-product operator; and $\lambda_s$ and $\lambda_g$ are weights controlling the contribution of the scale and rotation priors, respectively. These weights (\ie, $\lambda_s$ and $\lambda_g$) must be greater than or equal to zero.

The scale regularizer $\lambda_s \left( s_0 - s\right)^2$ imposes a penalty by deviating from the scale prior $s_0$. On the other hand, the rotation prior $\lambda_g \| \mathbf{g}_\mathcal{Q} \times R \mathbf{g}_\mathcal{W}\|^2$ imposes a misalignment penalty between the transformed world gravity direction $R \mathbf{g}_\mathcal{W}$ and the query gravity direction $\mathbf{g}_\mathcal{Q}$.

As discussed earlier, the first step to solve for pose and scale is to re-write the cost $J^\prime$ as a function that only depends on the rotation matrix. To do so, it is mathematically convenient to define
\begin{equation}
\mathbf{x} = \begin{bmatrix} \alpha_1 & \hdots & \alpha_n & s &\mathbf{t}^\intercal\end{bmatrix}^\intercal.
\end{equation}
The gradient evaluated at the optimal $\mathbf{x}^\star$ must satisfy the following constraint: $\nabla_{\mathbf{x}} J^\prime \bigr |_{\mathbf{x} = \mathbf{x}^\star} = 0$. From this constraint, we obtain the following relationship:
\begin{equation}
\begin{split}
\mathbf{x} &= \left(A^\intercal A  + P\right)^{-1} A^\intercal W \mathbf{b} + \left(A^\intercal A + P \right)^{-1} P\mathbf{x}_0 \\
 &= \begin{bmatrix} U \\ S \\ V\end{bmatrix} W \mathbf{b} + \lambda_s s_o \mathbf{l}
\end{split}
\label{eq:cost_as_rotation_fn}
\end{equation}
where 
\begin{equation}
\begin{split}
A &= \begin{bmatrix}  
\mathbf{r}_1 & & & \mathbf{c}_1 & -I \\
 & \ddots& & \vdots & \vdots \\
 & & \mathbf{r}_n & \mathbf{c}_n & -I 
\end{bmatrix}, 
\mathbf{b} = \begin{bmatrix} \mathbf{p}_1 \\ \vdots \\ \mathbf{p}_n \end{bmatrix} \\
P &= \begin{bmatrix} 
0_{n \times n} & & \\ 
  & \lambda_s & \\
  & & 0_{3 \times 3}
\end{bmatrix},
W = \begin{bmatrix} 
R & & \\
 & \ddots & \\
  & & R
\end{bmatrix}, 
\end{split}
\end{equation}
and $\mathbf{x}_0 = \begin{bmatrix} 0_{n}^\intercal & s_0 & 0_{3}^\intercal \end{bmatrix}^\intercal$. Inspired by gDLS~\cite{sweeney2014gdls} and DLS~\cite{hesch2011direct}, we partition $\left(A^\intercal A  + P\right)^{-1} A^\intercal$ into three matrices U, S, and V such that the depth, scale, and translation parameters are functions of U, S, and V, respectively. These matrices and the vector $\mathbf{l}$ can be computed in closed form by exploiting the sparse structure of the matrices $A$ and $P$; see appendix for the full derivation.

Eq.~\eqref{eq:cost_as_rotation_fn} provides a linear relationship between the depth, scale, and translation and the rotation matrix. Consequently, these parameters are computed as a function of the rotation matrix as follows:
\begin{equation}
\begin{split}
\alpha_i(R) &= \mathbf{u}_i^\intercal W \mathbf{b} + \lambda_s s_o \mathbf{l}_i \\
s(R) &= S W \mathbf{b} + \lambda_s s_o \mathbf{l}_{n + 1} \\
\mathbf{t}(R) &= V W \mathbf{b} + \lambda_s s_o \mathbf{l}_{\mathbf{t}},
\end{split}
\label{eq:rotation_relationships}
\end{equation} 
where $\mathbf{u}_i^\intercal$ is the $i$-th row of matrix $U$, $\mathbf{l}_j$ is the $j$-th entry of the vector $\mathbf{l}$, and $\mathbf{l}_{\mathbf{t}}$ corresponds to the last three entries of the vector $\mathbf{l}$.  Note that we can obtain the exact same relationships for depth, scale, and translation obtained by Sweeney~\etal for gDLS~\cite{sweeney2014gdls, sweeney2016large} when $\lambda_s = 0$. 

In order to re-write the regularized least-squares cost function (\ie, Eq.~\eqref{eq:regularized_scale_and_pose_problem}) as clearly as possible, we define 
\begin{equation}
\begin{split}
\mathbf{e}_i &= \alpha_i(R) \mathbf{r}_i - \left(R \mathbf{p}_i + \mathbf{t}(R) - s(R) \mathbf{c}_i \right) \\
 &= \boldsymbol{\eta}_i + \mathbf{k}_i \\
\boldsymbol{\eta}_i &= \mathbf{u}_i^\intercal W \mathbf{b} \mathbf{r}_i  - R \mathbf{p}_i  -VW\mathbf{b} + SW\mathbf{b} \mathbf{c}_i \\
\mathbf{k}_i &= \lambda_s s_0 \left( \mathbf{l}_i \mathbf{r} - \mathbf{l}_{\mathbf{t}} + \mathbf{l}_{n+1} \mathbf{c}_i \right).
\end{split}
\label{eq:residual_defs}
\end{equation}
The residual $\mathbf{e}_i$ is divided into two terms: $\boldsymbol{\eta}_i$, the residual part considering the unconstrained terms; and $\mathbf{k}_i$ the residual part considering the scale-prior-related terms. Note again that when $\lambda_s = 0$, $\mathbf{k}_i$ becomes null and $\mathbf{e}_i$ becomes the residual corresponding to gDLS~\cite{sweeney2014gdls, sweeney2016large}. 

Using the definitions from Eq.~\eqref{eq:residual_defs}, and the scale, depth, and translation relationships shown in Eq.~\eqref{eq:rotation_relationships}, we can now re-write the regularized least-squares cost function shown in Eq.~\eqref{eq:regularized_scale_and_pose_problem} as follows:
\begin{equation}
\begin{split}
J^\prime &= J^\prime_{\text{gDLS}} + J^\prime_s + J^\prime_{g} \\
 &= \text{vec}(R)^\intercal M \text{vec}(R) + 2 \mathbf{d}^\intercal \text{vec}(R) + k
 \label{eq:pgdls_quadratic_fn}
\end{split}
\end{equation}
where 
\begin{equation}
\begin{split}
J^\prime_{\text{gDLS}} &= \sum_{i=1}^n \mathbf{e}_i^\intercal \mathbf{e}_i = \sum_{i=1}^n \boldsymbol{\eta}_i^\intercal \boldsymbol{\eta}_i + 2 \mathbf{k}_i^\intercal \boldsymbol{\eta}_i + \mathbf{k}_i^\intercal \mathbf{k}_i \\
&= \text{vec}(R)^\intercal M_{\text{gDLS}} \text{vec}(R)  + 2\mathbf{d}^\intercal_{\text{gDLS}} \text{vec}(R) + k_{\text{gDLS}}\\
J^\prime_{s} &= \lambda_s \left( s_0 - S(R) \right)^2 \\
 &= \text{vec}(R)^\intercal M_s \text{vec}(R) + 2 \mathbf{d}^\intercal_s \text{vec}(R) + k_s \\
J^\prime_{g} &= \lambda_g \| \mathbf{g}_\mathcal{Q} \times R \mathbf{g}_\mathcal{W}\|^2 = \text{vec}(R)^\intercal  M_g \text{vec}(R) \\
M &= M_{\text{gDLS}} + M_s + M_g \\
\mathbf{d} &= \mathbf{d}_{\text{gDLS}} +  \mathbf{d}_s \\
k &= k_{\text{gDLS}} + k_s.
\end{split}
\end{equation}
The parameters of Eq.~\eqref{eq:pgdls_quadratic_fn} (\ie, $M_{\textbf{gDLS}}$, $M_s$, $M_g$, $\mathbf{d}_{\text{gDLS}}$, $\mathbf{d}_s$, $k_{\text{gDLS}}$, and $k_s$) can be computed in closed form and in $\mathcal{O}(n)$ time; see appendix for the closed form solutions of these parameters.

An important observation is that Eq.~\eqref{eq:pgdls_quadratic_fn} generalizes the unconstrained quadratic function of gDLS shown in Eq.~\eqref{eq:scale_and_pose_problem}. When both priors are disabled, \ie, $\lambda_g = \lambda_s = 0$, then $J^\prime(R) = J(R)$. Also, note that the weights $\lambda_g$ and $\lambda_s$ allow the user to control the contribution of each of the priors independently. This gives \MethodName~ great flexibility since it can be adapted to many scenarios. For instance, these weights can be adjusted so that \MethodName~ reflects the confidence on certain priors, reduces the effect of noise present in the priors, and fully disables one prior but enables another.

\begin{figure*}[t]
    \centering
    \includegraphics[width=\textwidth,height=9.5em]{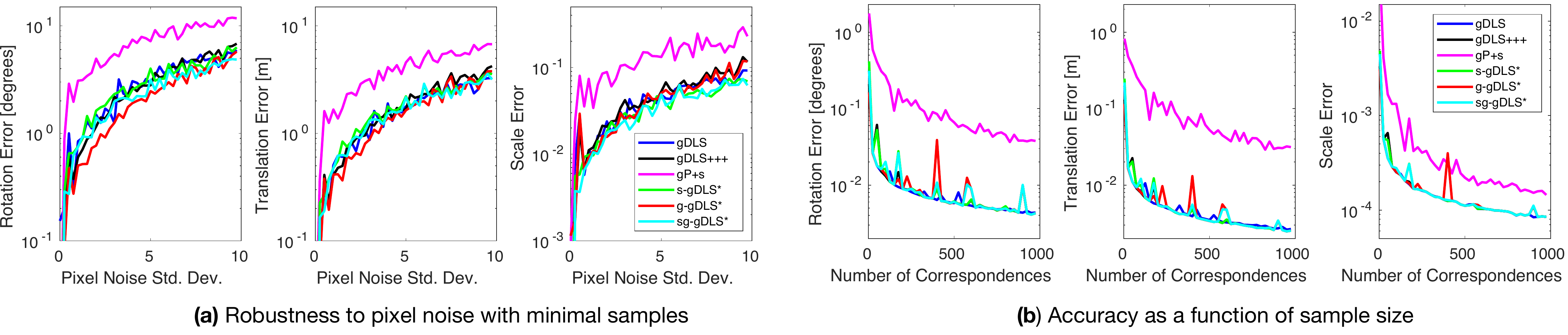}
    \caption{Rotation, translation, and scale errors as a function of {\bf (a)} pixel noise when estimating pose and scale using a minimal sample of correspondences, and {\bf (b)} sample size. s-\MethodName, g-\MethodName, and sg-\MethodName~ produce comparable errors to that of gDLS~\cite{sweeney2014gdls, sweeney2016large}. On the other hand, gP+s~\cite{ventura2014minimal} produces the highest errors.}
    \label{fig:robustness}
\end{figure*}

\subsection{Solving for Rotation}

Given that the prior-based pose-and-scale cost function (\ie, Eq.~\eqref{eq:regularized_scale_and_pose_problem}) depends only on the rotation matrix, the next step is to find $R$ such that it minimizes Eq.~\eqref{eq:pgdls_quadratic_fn}. To achieve this, \MethodName~ represents the rotation matrix $R$ using a quaternion $\mathbf{q} = \begin{bmatrix}
q_1 ~ q_2 ~ q_3 ~ q_4
\end{bmatrix} ^\intercal$. To compute all the minimizers of Eq.~\eqref{eq:pgdls_quadratic_fn}, \MethodName~ follows~\cite{hesch2011direct, kneip2014upnp, sweeney2014gdls, sweeney2016large} and builds a polynomial system that encodes the first-order optimality conditions and the unit-norm-quaternion constraint, \ie,
\begin{equation}
\begin{cases}
\frac{\partial J^\prime}{\partial q_j} = 0,  & \forall j=1, \hdots, 4 \\
q_j \left(\mathbf{q}^\intercal \mathbf{q} - 1 \right) = 0, & \forall j=1, \hdots, 4
\end{cases}.
\label{eq:poly_system}
\end{equation}
The polynomial system shown in Eq.~\eqref{eq:poly_system} encodes the unit-norm-quaternion constraint with 
\begin{equation}
\frac{\partial \left( \mathbf{q}^\intercal \mathbf{q} - 1 \right)^2}{\partial q_j} = q_j \left(\mathbf{q}^\intercal \mathbf{q} - 1 \right) = 0, \forall j.
\label{eq:unit_norm_quaternion_constraint}
\end{equation} 
Eq.~\eqref{eq:unit_norm_quaternion_constraint} yields efficient elimination templates and small action matrices, which delivers efficient polynomial solvers as Kneip~\etal~\cite{kneip2014upnp} shows. In fact, \MethodName~ adopts the efficient polynomial solver of Kneip~\etal~\cite{kneip2014upnp} as we leverage their rotation representation
\begin{equation}
\text{vec}(R) = \left[
q_1^2 ~ q_2^2 ~ q_3^2 ~ q_4^2 ~ q_1 q_2 ~ q_1 q_3 ~ q_1 q_4 ~ q_2 q_3 ~ q_2 q_4 ~ q_3 q_4 \right]^\intercal.
\end{equation}
Given this representation, the dimensions of the parameters of the regularized least-squares cost function shown in Eq.~\eqref{eq:pgdls_quadratic_fn} become $M \in \mathbb{R}^{10 \times 10}$, $\mathbf{d} \in \mathbb{R}^{10}$, and $k \in \mathbb{R}$.  

Because \MethodName~ uses the solver of Kneip~\etal~\cite{kneip2014upnp}, it efficiently computes eight rotations. After computing these solutions, \MethodName~ discards quaternions with complex numbers, and then recovers the depth, scale, and translation using Eq.~\eqref{eq:rotation_relationships}. Finally, \MethodName~ uses the computed similarity transformations to discard solutions that map the input 3D points behind the camera.

{\bf Our gDLS* derivation can be generalized. }Imposing scale and translation priors via the regularizers is general enough to be adopted by least-squares-based estimators (\eg, DLS~\cite{hesch2011direct} and UPnP~\cite{kneip2014upnp}). This is because the regularizers are quadratic functions that can be added without much effort into their derivations.

%% file: tex/Results.tex
\section{Experiments}
\label{sec:results}

\begin{figure}[t]
    \centering
    \includegraphics[width=\columnwidth]{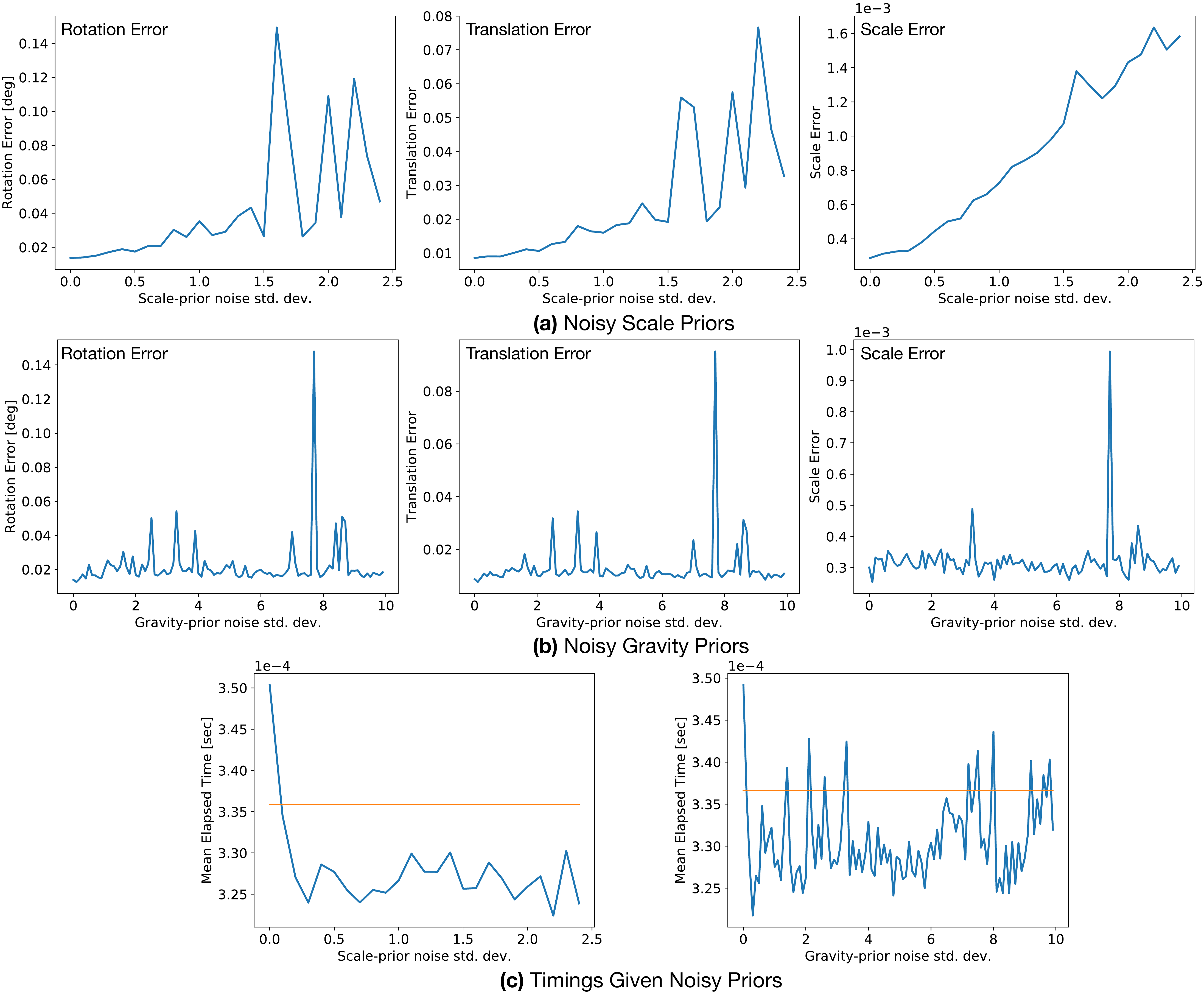}
    \caption{{\bf (a)} \MethodName's accuracy slowly degrades as the noise increases in the scale prior. {\bf (b)} \MethodName's accuracy is barely affected by noise in the gravity prior. {\bf (c)} Time is not affected when using noisy scale (left) and gravity (right) priors. The orange line shows the best timing of gDLS+++~\cite{sweeney2016large}, the second fastest estimator in our experiments.}
    \label{fig:noisy_priors}
\end{figure}

\begin{table*}[t]
    \centering
    \caption{Estimation times in seconds for gP+s~\cite{ventura2014minimal}, gDLS~\cite{sweeney2014gdls}, gDLS+++~\cite{sweeney2016large}, UPnP~\cite{kneip2014upnp}, s-\MethodName~(s column), g-\MethodName~(g column), and sg-\MethodName~(sg column) for rigid ($s_0=1$) and similarity ($s_0=2.5$) transformations. The top six rows show results for the TUM dataset, and the last six rows show results for the KITTI dataset. A gravity prior tends to deliver fast estimates, while a scale prior modestly slows down the estimation. On the other hand, scale and gravity priors tend to be modestly faster than gDLS+++.}
    \footnotesize {
    \begin{tabular}{l ccccccc c cccccc}
        \toprule
        & \multicolumn{7}{c}{Rigid Transformation $\left[s_0=1\right]$} & & \multicolumn{6}{c}{Similarity Transformation $\left[s_0=2.5\right]$}\\
        \cmidrule{2-8} \cmidrule{10-15}
        &
        \cite{ventura2014minimal} & \cite{sweeney2014gdls} & \cite{sweeney2016large} & \cite{kneip2014upnp} & 
        s & g & sg & &
        \cite{ventura2014minimal} & \cite{sweeney2014gdls} & \cite{sweeney2016large} & s & g & sg\\
        \cmidrule{1-15}
        Fr1 Desk & 
        10.77 & 15.38 & \zp{5.79} & 9.60 & 8.11 & \fp{3.39} & 5.80 & &
        10.54 & 15.38 & \zp{5.79} & 8.12 & \fp{3.82} & 6.89 \\
        Fr1 Room & 
        8.86 & 12.23 & \zp{4.51} & 5.52 & 6.66 & \fp{2.85} & 4.70 & &
        8.55 & 11.98 & \zp{4.61} & 7.96 & \fp{3.24} & 5.40 \\
        Fr2 LargeNoLoop & 
        5.63 & 8.10 & 2.76 & 2.23 & 2.52 & \fp{2.01} & \zp{2.10} & &
        6.46 & 8.25 & 2.82 & 2.49 & \zp{2.40} & \fp{2.39} \\
        Fr1 Desk2 & 
        4.99 & 7.72 & \zp{2.96} & 4.20 & 3.89 & \fp{1.67} & 3.65 & &
        5.04 & 7.47 & \zp{2.59} & 3.99 & \fp{1.86} & 3.38 \\
        Fr2 Pioneer SLAM & 
        21.06 & 27.77 & 10.82 & 8.27 & 12.08 & \fp{6.48} & \zp{8.08} & &
        17.10 & 25.42 & \zp{9.34} & 11.84 & \fp{7.41} & 9.70\\
        Fr2 Pioneer SLAM 2 & 
        3.39 & 3.49 & 1.93 & \fp{0.88} & 1.17 & \zp{1.02} & 1.40 & &
        3.16 & 3.66 & 1.96 & 1.17 & \zp{1.09} & \fp{0.99} \\
        \cmidrule{1-15}
        Drive 1 $\left( 10^{-2} ~ [\text{sec}] \right)$& 
        1.66 & 2.32 & 0.93 & 1.17 & \zp{0.90} & \fp{0.58} & \zp{0.90} & &
        1.31 & 1.79 & 1.19 & 1.0 & \fp{0.61} & \zp{0.87} \\
        Drive 9 & 
        0.32 & 0.49 & 0.23 & \fp{0.14} & 0.34 & \zp{0.23} & 0.28 & &
        0.35 & 0.48 & \fp{0.22} & 0.50 & \fp{0.22} & 0.26 \\
        Drive 19 & 
        0.51 & 0.29 & 0.80 & \zp{0.27} & 0.29 & \fp{0.26} & 0.29 & &
        0.57 & 0.90 & 0.39 & \zp{0.29} & \fp{0.26} & \zp{0.29}\\
        Drive 22 & 
        0.12 & 0.22 & 0.12 & \zp{0.07} & 0.12 & \fp{0.06} & 0.11 & &
        0.12 & 0.19 & \zp{0.07} & 0.11 & \fp{0.06} & 0.12 \\
        Drive 23 $\left( 10^{-2} ~ [\text{sec}] \right)$& 
        3.40 & 4.72 & 2.14 & 2.18 & 2.19 & \fp{1.45} & \zp{2.04} & &
        3.25 & 4.71 & 2.07 & 2.34 & \fp{1.55} & \zp{1.94}\\
        Drive 29 & 
        1.15 & 1.95 & \fp{0.76} & 1.19 & 1.13 & \zp{0.77} & 1.11 & &
        1.16 & 1.96 & \zp{0.75} & 1.14 & \fp{0.74} & 1.12\\
        \bottomrule
    \end{tabular}
    }
    \label{tab:times}
\end{table*}

This section presents experiments that use (i) synthetic data to demonstrate the numerical stability and robustness of \MethodName~and (ii) real data to show the performance of \MethodName~in registering a SLAM trajectory to a pre-computed point cloud. We test three \MethodName~configurations: scale-only-regularized (s-\MethodName), gravity-only-regularized (g-\MethodName), and scale-gravity-regularized (sg-\MethodName). For all experiments except the ablation study, $\lambda_s$ and $\lambda_g$ are fixed to $1$. We compare to several state-of-the-art pose-and-scale estimators: gP+s~\cite{ventura2014minimal}, gDLS~\cite{sweeney2014gdls}, gDLS+++~\cite{sweeney2016large}, and UPnP~\cite{kneip2014upnp}. All implementations are integrated into Theia-SfM~\cite{sweeney2015theia}. For all experiments, we use one machine with two 2.10~GHz Intel Xeon CPUs and 32 GB of RAM.


{\bf Datasets.} For the SLAM trajectory registration, the experiments use two publicly available SLAM datasets: the TUM RGBD dataset~\cite{sturm12iros} and the KITTI dataset~\cite{Geiger2013IJRR}. These datasets provide per-frame accelerometer estimates, which we use to compute one gravity direction for each SLAM trajectory. Specifically, we low pass filter and smooth the accelerations (because the gravity acceleration is constant within the high frequency noise) to get an estimate of the gravity vector for each image. Then, we take the mean of all these estimates to get a final gravity vector estimate. The final result is one gravity vector for each trajectory.

{\bf Error Metrics.} All the experiments report rotation, translation, and scale errors. The rotation error is the angular distance~\cite{hartley2013rotation, huynh2009metrics} between the expected and the estimated rotation matrix. The translation error is the L2 norm between the expected and the estimated translation. Lastly, the scale error is the absolute difference between the expected and the estimated scale values.




\subsection{Robustness to Noisy Synthetic Data}
\label{sec:robustness}

This experiment consists of three parts: (1) measuring robustness to pixel noise with minimal samples (\ie, four 2D-3D correspondences); (2) measuring accuracy as a function of the size of a non-minimal sample (\ie, more than four 2D-3D correspondences); and (3) testing how noise in scale and gravity priors effects solution accuracy and run time. For all experiments, we execute $1,000$ trials using $10$ randomly positioned cameras within the cube $\left[-10, 10\right] \times \left[-10, 10\right] \times \left[-10, 10 \right]$, and $300$ random 3D points in the cube $\left[-5, 5\right] \times \left[-5, 5\right] \times \left[10, 20 \right]$. For each trial, we transform the 3D points by the inverse of a randomly generated ground truth similarity transformation (i.e., a random unit vector direction and random rotation angle between $0^{\circ}$ and $360^{\circ}$, random translation between $0$ and $5$ in $(x,y,z)$, and random scale between $0$ and $5$).

\begin{figure}[t]
    \centering
    \includegraphics[width=0.48\textwidth]{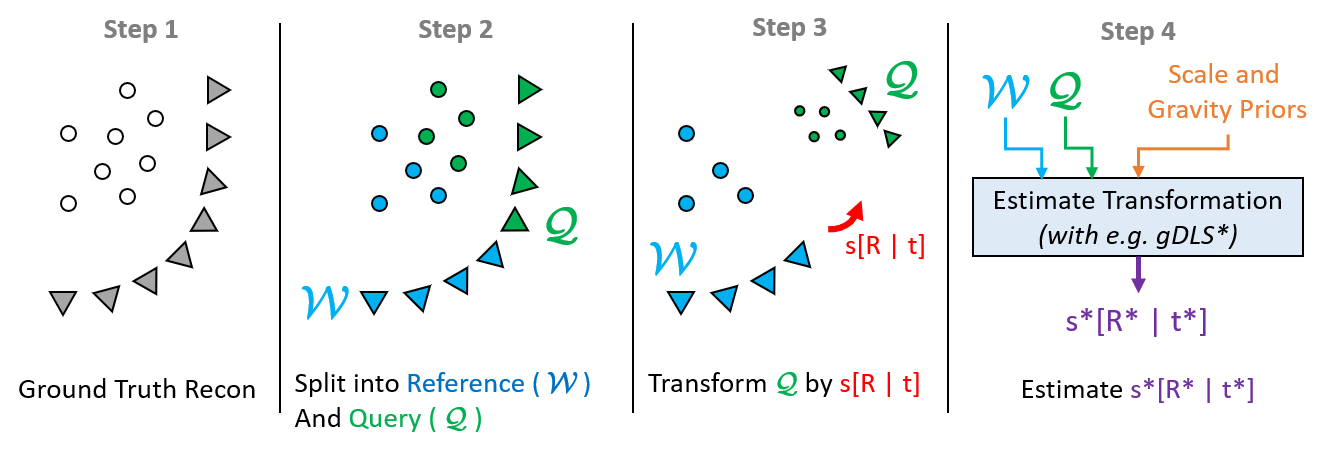}
    \caption{Evaluation Protocol: (1) Reconstruct a scene using Theia-SfM; (2) Split the reconstruction into Query ($\mathcal{Q}$) and Reference ($\mathcal{W}$) parts; (3) Transform $\mathcal{Q}$ using a similarity transform $S$; and (4) Estimate similarity transform $S^\star$ aligning $\mathcal{Q}$ and $\mathcal{W}$.}
    \label{fig:experiment_setup}
\end{figure}

{\bf \MethodName~ is robust to pixel noise with minimal samples.} We generate random minimal samples with zero-mean Gaussian noise added to the pixel positions. We vary the noise standard deviation between $0$ and $10$ and measure the rotation, translation, and scale errors. The results of this experiment can be seen in Fig.~\ref{fig:robustness}(a). We observe that s-\MethodName, g-\MethodName, and sg-\MethodName~ perform similarly to gDLS~\cite{sweeney2014gdls} and gDLS+++~\cite{sweeney2016large} when comparing their rotation and translation errors. Also, we see that the rotation and translation errors produced by gP+s~\cite{ventura2014minimal} are the highest. All methods produce similar scale errors.

{\bf Accuracy of \MethodName~ improves with non-minimal samples.} For this experiment, we vary the size of the sample from $5$ to $1000$ 2D-3D correspondences and fix the standard deviation to $0.5$ for the zero-mean Gaussian pixel noise. Fig.~\ref{fig:robustness}(b) shows that s-\MethodName, g-\MethodName, and sg-\MethodName~ produce comparable rotation, translation, and scale errors to that of gDLS and gDLS+++. On the other hand, gP+s produced the largest errors.

\begin{table*}[t]
    \centering
    \caption{Rotation, translation, and scale errors of gP+s~\cite{ventura2014minimal}, gDLS~\cite{sweeney2014gdls}, gDLS+++~\cite{sweeney2016large}, UPnP~\cite{kneip2014upnp}, and \MethodName~ using a unit scale (\ie, $s_0 = 1$) and gravity priors. The first six rows show results for the TUM dataset, and the last six rows show results for the KITTI dataset. The smallest errors are shown in bold. We observe that \MethodName~ and UPnP perform equivalently when comparing rotation and translation errors. However, \MethodName~ produces the lowest errors among the pose-and-scale estimators (\ie, gP+s, gDLS, and gDLS+++).}
    \footnotesize {
    \begin{tabular}{l ccccc c ccccc c cccc}
        \toprule
        & \multicolumn{15}{c}{Rigid Transformation $\left[s_0=1\right]$} \\
        \cmidrule{1-17}
        & \multicolumn{5}{c}{$R_{\text{error}} \left[\text{deg}\right] ~ (10^{-1})$} & & \multicolumn{5}{c}{$\mathbf{t}_\text{error} ~ (10^{-2})$} & & \multicolumn{4}{c}{$s_{\text{error}} ~ (10^{-3})$} \\
        \cmidrule{2-6} \cmidrule{8-12} \cmidrule{14-17}
        &
        \cite{ventura2014minimal} & \cite{sweeney2014gdls} & \cite{sweeney2016large} & \cite{kneip2014upnp} & Ours~ & &
        \cite{ventura2014minimal} & \cite{sweeney2014gdls} & \cite{sweeney2016large} & \cite{kneip2014upnp} & Ours~ & &
        \cite{ventura2014minimal} & \cite{sweeney2014gdls} & \cite{sweeney2016large} & Ours~\\
        \cmidrule{1-17}
        Fr1 Desk & 
        2.78 & 2.58 & 2.72 & 1.89 & \fp{1.57} & &
        2.02 & 1.92 & 1.91 & 1.36 & \fp{1.19} & &
        1.46 & 1.33 & 1.29 & \fp{0.68}\\
        Fr1 Room & 
        2.05 & 1.95 & 1.99 & 1.05 & \fp{0.98} & &
        1.09 & 1.05 & 1.09 & 0.55 & \fp{0.52} & &
        1.48 & 1.40 & 1.39 &  \fp{0.32}\\
        Fr2 LargeNoLoop & 
        1.90 & 1.61 & 1.62 & 1.35 & \fp{1.32} & &
        6.77 & 5.77 & 6.01 & 4.70 & \fp{4.45} & &
        4.04 & 4.15 & 4.39 & \fp{1.39}\\
        Fr1 Desk2 & 
        2.34 & 2.13 & 2.26 & 1.63 & \fp{1.25} & &
        2.03 & 1.85 & 1.90 & 1.38 & \fp{1.06} & &
        1.35 & 1.15 & 1.26 &  \fp{0.49} \\
        Fr2 Pioneer SLAM & 
        1.50 & 1.51 & 1.47 & \fp{0.76} & 0.87 & &
        1.29 & 1.18 & 1.19 & \fp{0.59} & 0.70 & &
        2.77 & 2.73 & 2.67 &  \fp{0.87}\\
        Fr2 Pioneer SLAM 2& 
        1.63 & 1.49 & 1.51 & \fp{0.97} & 1.08 & &
        1.93 & 1.77 & 1.80 & \fp{1.22} & 1.34 & &
        6.81 & 7.48 & 7.44 & \fp{0.88}\\
        \cmidrule{1-17}
        Drive 1 & 
        0.44 & 0.40 & 0.42 & 0.34 & \fp{0.33} & &
        0.27 & 0.25 & 0.24 & 0.17 & \fp{0.16} & &
        0.72 & 0.66 & 0.61 & \fp{0.02} \\
        Drive 9 & 
        1.15 & 1.10 & 1.15 & \fp{0.64} & 1.13 & &
        0.43 & 0.40 & 0.44 & \fp{0.13} & 0.20 & &
        6.27 & 5.79 & 6.14 & \fp{0.05}\\
        Drive 19 & 
        3.42 & 3.57 & 3.30 & \fp{2.48} & 3.04 & &
        0.83 & 0.85 & 0.80 & \fp{0.63} & 0.73 & &
        4.99 & 5.64 & 5.54 & \fp{0.01}\\
        Drive 22 & 
        0.66 & 0.62 & 0.66 & \fp{0.31} & 0.63 & &
        0.28 & 0.27 & 0.30 & \fp{0.16} & 0.27 & &
        1.90 & 1.70 & 1.67 & \fp{0.94}\\
        Drive 23 & 
        0.74 & 0.58 & 0.62 & 0.84 & \fp{0.56} & &
        0.19 & 0.18 & 0.19 & 0.12 & \fp{0.09} & &
        1.28 & 1.16 & 1.28 & \fp{0.03}\\
        Drive 29 & 
        1.00 & 1.06 & 1.06 & \fp{0.56} & 0.82 & &
        0.34 & 0.35 & 0.35 & \fp{0.21} & 0.26 & &
        1.60 & 3.39 & 1.73 & \fp{0.75}\\
        \bottomrule
    \end{tabular}
    }
    \label{tab:rigid_results}
\end{table*}

\begin{table*}[t]
    \centering
    \caption{Rotation, translation, and scale errors of gP+s~\cite{ventura2014minimal}, gDLS~\cite{sweeney2014gdls}, gDLS+++~\cite{sweeney2016large}, and \MethodName~ using a scale prior of $s_0 = 2.5$ and gravity priors. The first six rows show results for the TUM dataset, and the last six rows show results for the KITTI dataset. The smallest errors are shown in bold. We see that \MethodName~ produces the smallest errors in almost every case.}
    \footnotesize {
    \begin{tabular}{l cccc c cccc c cccc}
        \toprule
        & \multicolumn{14}{c}{Similarity Transformation $\left[s_0=2.5\right]$} \\
        \cmidrule{1-15}
        & \multicolumn{4}{c}{$R_{\text{error}} \left[\text{deg}\right] ~ (10^{-1})$} & & \multicolumn{4}{c}{$\mathbf{t}_\text{error} ~ (10^{-2})$} & & \multicolumn{4}{c}{$s_{\text{error}} ~ \left(10^{-3}\right)$} \\
        \cmidrule{2-5} \cmidrule{7-10} \cmidrule{12-15}
        &
        \cite{ventura2014minimal} & \cite{sweeney2014gdls} & \cite{sweeney2016large} & Ours~ & &
        \cite{ventura2014minimal} & \cite{sweeney2014gdls} & \cite{sweeney2016large} & Ours~ & &
        \cite{ventura2014minimal} & \cite{sweeney2014gdls} & \cite{sweeney2016large} & Ours~\\
        \cmidrule{1-15}
        Fr1 Desk & 
        2.77 & 2.59 & 2.72 & \fp{2.23} & &
        5.50 & 5.21 & 5.22 & \fp{4.55} & &
        3.59 & 3.27 & 3.23 & \fp{1.56} \\
        Fr1 Room & 
        2.02 & 1.99 & 1.99 & \fp{1.29} & &
        2.79 & 2.78 & 2.79 & \fp{1.71} & &
        3.74 & 3.41 & 3.48 & \fp{0.91}\\
        Fr2 LargeNoLoop & 
        1.90 & 1.57 & 1.62 & \fp{1.49} & &
        13.9 & 12.3 & 12.8 & \fp{10.1} & &
        10.7 & 10.8 & 11.0 & \fp{3.66} \\
        Fr1 Desk2 & 
        2.29 & 2.13 & 2.26 & \fp{1.77} & &
        4.37 & 4.14 & 4.32 & \fp{3.28} & &
        3.22 & 2.88 & 3.14 & \fp{1.37}\\
        Fr2 Pioneer SLAM & 
        1.43 & 1.48 & 1.47 & \fp{1.17} & &
        3.67 & 3.44 & 3.50 & \fp{2.42} & &
        6.98 & 6.81 & 6.68 & \fp{1.65}\\
        Fr2 Pioneer SLAM 2& 
        1.84 & 1.49 & 1.51 & \fp{1.27} & &
        5.23 & 4.49 & 4.58 & \fp{3.69} & &
        15.8 & 18.7 & 18.6 & \fp{2.42}\\
        \cmidrule{1-15}
        Drive 1 & 
        0.43 & 0.40 & 0.42 & \fp{0.33} & &
        0.47 & 0.44 & 0.43 & \fp{0.28} & &
        1.73 & 1.62 & 1.54 & \fp{0.05}\\
        Drive 9 & 
        1.17 & \fp{1.09} & 1.15 & 1.13 & &
        1.11 & 1.06 & 1.16 & \fp{0.50} & &
        15.7 & 14.4 & 15.3 & \fp{0.12}\\
        Drive 19 & 
        3.60 & 3.27 & 3.57 & \fp{3.09} & &
        2.13 & 1.91 & 2.06 & \fp{1.80} & &
        14.6 & 14.0 & 14.1 & \fp{0.04} \\
        Drive 22 & 
        0.64 & \fp{0.62} & 0.66 & \fp{0.62} & &
        0.77 & \fp{0.75} & 0.82 & \fp{0.75} & &
        4.40 & 4.27 & 4.19 & \fp{2.34}\\
        Drive 23 & 
        0.75 & 0.59 & 0.62 & \fp{0.57} & &
        0.58 & 0.51 & 0.56 & \fp{0.24} & &
        3.13 & 2.87 & 3.20 & \fp{0.14}\\
        Drive 29 & 
        1.00 & 1.09 & 1.06 & \fp{0.82} & &
        0.80 & 0.89 & 0.88 & \fp{0.65} & &
        3.81 & 4.46 & 4.33 & \fp{1.86}\\
        \bottomrule
    \end{tabular}
    }
    \label{tab:sim_results}
\end{table*}

{\bf \MethodName~ is numerically stable. }From Fig.~\ref{fig:robustness}, we conclude that s-\MethodName~(green), g-\MethodName~(red), sg-\MethodName~(cyan) are numerically stable because the errors are similar to that of gDLS+++.

{\bf \MethodName~is robust to noise in scale and gravity priors. }For this experiment, we gradually increase the noise in the scale and gravity priors. In Fig.~\ref{fig:noisy_priors}(a), we see that noise in the scale prior slowly increases the rotation, translation, and scale errors. Conversely, in Fig.~\ref{fig:noisy_priors}(b), noise in the gravity prior has little effect on the final accuracy. Lastly, Fig.~\ref{fig:noisy_priors}(c) shows that noise has a minimal effect on the solution time.

\subsection{SLAM Trajectory Registration}
\label{sec:slam_registration}

The goal of this experiment is to measure the accuracy of an estimated similarity transformation which registers a SLAM trajectory (a collection of images from a moving camera) to a pre-computed 3D reconstruction. This experiment uses both scale and gravity priors for \MethodName. Part of this experiment considers a unit-scale similarity transformation, which makes it equivalent to a rigid transformation. In the latter case, the experiment also includes UPnP~\cite{kneip2014upnp}, a state-of-the-art multi-camera pose estimator that only estimates a rigid transformation (\ie, no scale estimation). 

\begin{figure*}[t]
    \centering
    \includegraphics[width=\textwidth,height=8.5em]{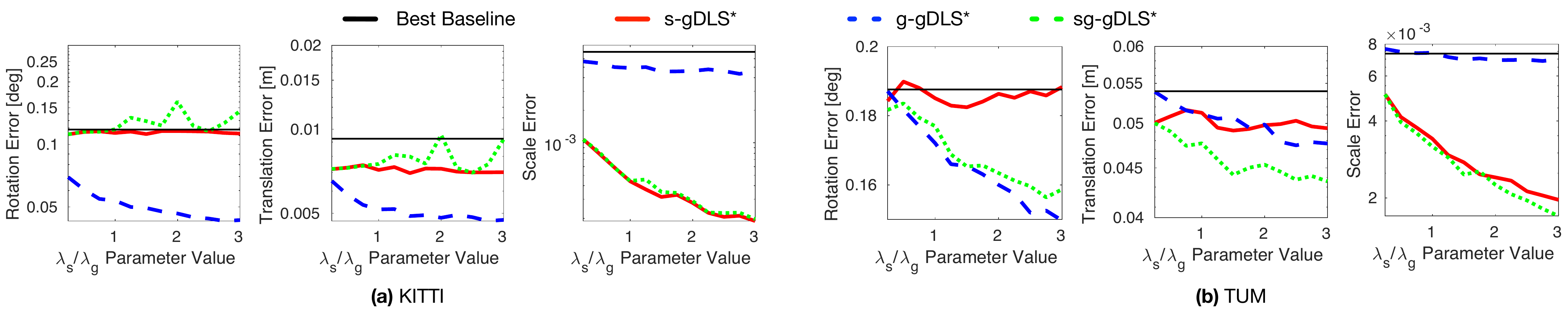}
    \caption{Average rotation, translation, and scale errors of the best baseline (best performing baseline for a given metric) and \MethodName~ as a function of $\lambda_s$ and $\lambda_g$ on {\bf(a)} KITTI and {\bf(b)} TUM datasets. A gravity prior (g-\MethodName) tends to reduce rotation and translation errors and modestly improves scale errors. A scale prior (s-\MethodName) tends to improve scale accuracy and modestly reduces translation errors. The combination of scale and gravity priors (sg-\MethodName) tends to reduce translation and scale errors and improves rotation estimates (see {\bf(b)}).}
    \label{fig:ablation_fig}
\end{figure*}

For each dataset and method combination, we run $100$ trials. Each estimator is wrapped in RANSAC~\cite{fischler1981random} to estimate the transformations and the same parameters are used for all of the scale-and-pose experiments. RANSAC labels correspondences with more than $4$ pixels of reprojection error as outliers. Because we use RANSAC, all methods tend to converge to accurate solutions (Tables~\ref{tab:rigid_results} and \ref{tab:sim_results}); however, the speed of convergence can differ significantly (Table~\ref{tab:times}).

While there exist datasets and clear methods to evaluate visual-based localization or SfM reconstructions (\eg, \cite{degol2018improved, sattler2018benchmarking}), there is not a well established methodology to evaluate pose-and-scale estimators. Previous evaluation procedures (\eg, \cite{sweeney2014gdls, sweeney2016large}) mostly show camera position errors, but discard orientation and scale errors. To evaluate the registration of a SLAM trajectory, we propose a novel evaluation procedure as illustrated in Fig.~\ref{fig:experiment_setup}: (1) reconstruct the trajectory using Theia-SfM; (2) remove a subset of images with their corresponding 3D points and tracks to create a new query set (the remaining images, points, and tracks are the reference reconstruction); (3) apply a similarity transformation to describe the reconstruction in a different frame of reference with a different scale; and (4) estimate the similarity transformation. To compute the input 2D-3D correspondences, the evaluation procedure matches the features from the query images to the features of the reference reconstruction and geometrically verifies them. From these matches and reconstruction, the procedure builds the 2D-3D correspondences by first computing the rays pointing to the corresponding 3D points using the camera positions.

{\bf The gravity prior significantly improves speed.} Table~\ref{tab:times} shows the average estimation times for both rigid ($s_0=1$) and similarity ($s_0=2.5$) transformations. We observe that both priors help the estimators find the solution much faster than many baselines (see sg columns). In particular, a gravity only prior (see g columns) can speed up \MethodName~ significantly while producing good estimates (see Sec.~\ref{sec:ablation_study}). On the other hand, a scale only prior (see s columns) can modestly accelerate \MethodName.

{\bf Incorporating scale and rotation priors consistently improves accuracy.} Tables~\ref{tab:rigid_results} and~\ref{tab:sim_results} present the average rotation, translation, and scale errors of $100$ trials, each estimating rigid and similarity transformations, respectively. Both Tables show six TUM trajectories at the top and six KITTI trajectories at the bottom. The scale priors $s_0$ are shown at the top of both Tables. Note that UPnP does not estimate scales, so it is not included in similarity transformation sections. Table~\ref{tab:rigid_results} shows that \MethodName~ and UPnP produce the most accurate rotation and translation estimates, and that \MethodName~ produces the most accurate scale estimates. Table~\ref{tab:sim_results} shows that \MethodName~ tends to produce the most accurate rotation and translation estimates, and that \MethodName~ produces the most accurate scale estimates.

\subsection{Ablation Study}
\label{sec:ablation_study}

This study aims to show the impact on the estimator accuracy of the weights $\lambda_s$ and $\lambda_g$ as they vary. We use the same TUM and KITTI datasets and RANSAC configuration as in previous experiments. We vary the weights from $0.25$ to $3$ using increments of $0.25$ and run $100$ trials for each weight. To summarize the results, we average the rotation, translation, and scale errors. 

{\bf The priors improve accuracy and speed when used individually or together.} Fig.~\ref{fig:ablation_fig} shows the results of this study. We see that on average a gravity prior (g-\MethodName) significantly improves rotation and translation errors, while modestly improving scale errors. On the other hand, a scale prior (s-\MethodName) on average significantly improves the scale errors, while modestly improving translation errors. Finally, both gravity and scale priors improve translation and scale errors and can help the estimator improve rotation errors.
    
From these results, we can conclude that accurate priors can greatly improve accuracy estimates (thereby also improving speed). However, we know from Fig.~\ref{fig:noisy_priors} that noisy priors can also degrade accuracy. Thus, for future work, we will explore how to automatically set $\lambda_s$ and $\lambda_g$ based on the noise of the priors to maximize accuracy and speed.

%% file: tex/Conclusion.tex
\section{Conclusion}
\label{sec:conclusion}

This work presents \MethodName, a novel pose-and-scale estimator that exploits scale and/or gravity priors to improve accuracy and speed. \MethodName~ is based on a least-squares re-projection error cost function which facilitates the use of regularizers that impose prior knowledge about the solution space. 
This gDLS* derivation is general because these regularizers are quadratic functions that can easily be added to other least-squares-based estimators.
Experiments on both synthetic and real data show that \MethodName~ improves speed and accuracy of the pose-and-scale estimates given sufficiently accurate priors. The gravity prior is particularly effective, but the scale prior also improves the translation and scale estimates. 
These findings make \MethodName~ an excellent estimator for many applications where inertial sensors are available such as AR, 3D mapping, and robotics.